\documentclass[biblatex,a4paper]{svproc}
\usepackage[hyphens]{url}
\usepackage{microtype}
\usepackage{url}
\usepackage{cleveref}
\usepackage{graphicx}
\usepackage{tikz}
\usepackage{pgfplots}
\pgfplotsset{compat=1.18}
\usepackage[backend=biber,style=numeric-comp,isbn=false,url=false,eprint=false,maxbibnames=7,doi=false,giveninits=true]{biblatex}
\AtEveryBibitem{\clearfield{pages}} 
\AtEveryBibitem{\clearlist{language}}

\begin{document}
\mainmatter              
\title{Measuring and Minimizing Disturbance of Marine Animals to Underwater Vehicles}
\titlerunning{Animal Disturbance Monitoring with AUVs}  
%
\author{Levi Cai\inst{1} \and Youenn Jezequel\inst{2} \and
T. Aran Mooney\inst{2} \and Yogesh Girdhar\inst{2}}
%
%
%
\institute{Massachusetts Institute of Technology and Woods Hole Oceanographic Institution Joint Program\\
\email{cail@mit.edu},\\ 
\and
Woods Hole Oceanographic Institution}

\maketitle              

\begin{abstract}
    Do fish respond to the presence of underwater vehicles, potentially biasing our estimates about them? If so, are there strategies to measure and mitigate this response? This work provides a theoretical and practical framework towards bias-free estimation of animal behavior from underwater vehicle observations. We also provide preliminary results from the field in coral reef environments to address these questions. 
\end{abstract}


%
\section{Introduction}
\label{sec:intro}

Autonomous underwater vehicles (AUVs) and remotely operated vehicles (ROVs) have been proposed as scalable alternatives or supplementary systems for diver-based surveys and trawling methods to estimate species abundance, biodiversity, and animal behavior in a variety of environments ranging from coral reefs to deeper benthic communities \cite{clarke_using_2009,obura_coral_2019}. Many of these approaches rely on cameras to record videos or imagery during vehicle surveys, which can then be used to visually count animals or study behavior directly.

Recent studies of marine animal interactions with underwater vehicles \cite{kruusmaa_salmon_2020,maxeiner_social_2023} suggest that the presence of vehicles themselves can lead to a bias in the estimation of these quantities, impacting scientists' ability to trust those measurements. For instance, if animals are repelled or attracted to a vehicle, this will cause under- or over-estimation of their populations, respectively. However, many of these studies were conducted on animals \textit{in captivity}, which often do not reflect the behaviors of animals in the wild \cite{turko_physiological_2023}. 

Campbell et al. \cite{campbell_assessment_2021} provides one of the first studies of marine animal-robot interaction in the wild. They used a camera trap setup and measured fish responses to AUVs, ROVs, and Towed Vehicles (TVs). Benoit-Bird et al. \cite{benoit-bird_schrodingers_2023} followed with a study about deep sea fish response to light on AUVs and ROVs by using echosounders for measurements. In both studies, their initial findings suggest that different vehicles do produce noticeable shifts in fish counts. However, due to the logistical challenges of each setup, it is difficult to reproduce and validate a variety of other hypotheses in other types of environments or species. The goals of this work is to:
\begin{enumerate}
    \item Provide a theoretical framework to identify and \textit{systematically} minimize measurement bias of AUVs for collecting data about marine animal behavior, that is agnostic to sensor and emittance types, which allows us to methodically test across them.
    \item Provide and validate a simple protocol for collecting data \textit{in-situ} to inform the model.
\end{enumerate}

We propose a simplified framework towards bias-free estimation of animal behavior. In these studies, we focus primarily on coral reef fish species that exhibit \textit{hiding} behaviors, but we believe this framework could be extended to support other species, behaviors, and environments. It relies on an AUV/ROV and external stationary cameras to establish control and disturbance behaviors. We present initial results from the field suggesting that some common species of coral reef organisms exhibit responses to the presence of vehicles, which requires further study of the utility of AUVs for these purposes. We also discuss how these results can be integrated into vehicle control systems to potentially mitigate these impacts.

\section{Technical Approach and Methods}
\label{sec:approach}

\subsection{Model of animal response}

\begin{figure}
    \centering
    \includegraphics[width=\linewidth]{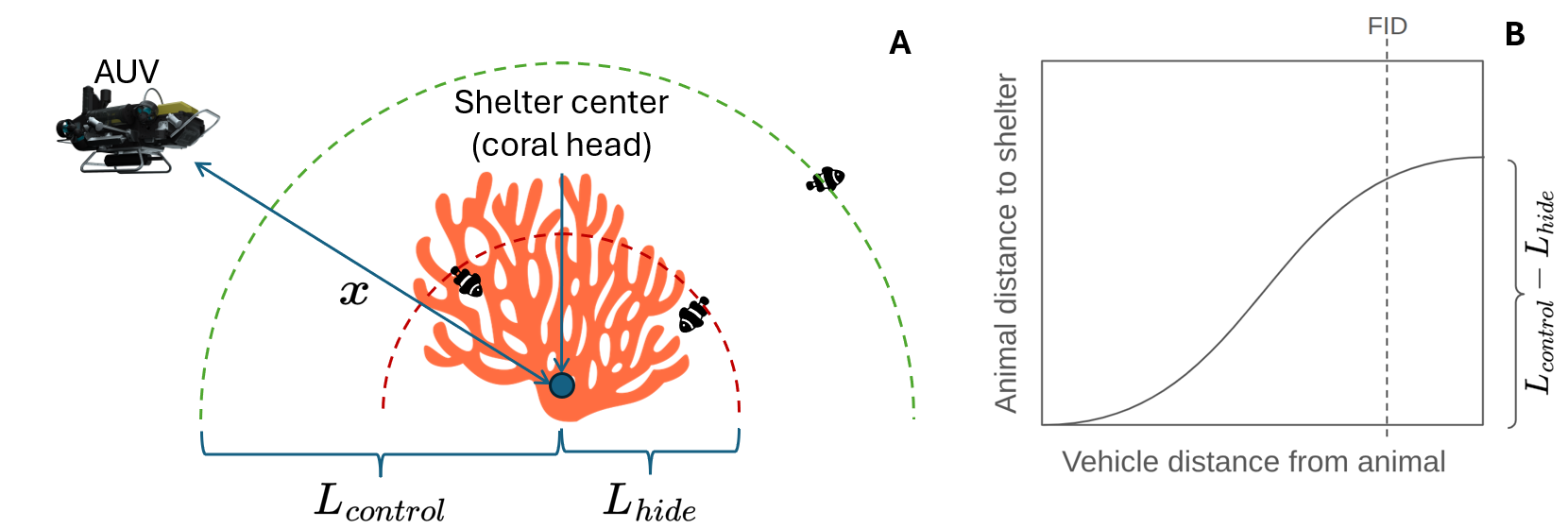}
    \caption[response-qual-model]{Qualitative overview of our model of fish response to AUVs, as some fish species are known to hide once a predator reaches a flight initiation distance (FID). FID can be estimated from this approach, modelled in \Cref{eqn:model}, as shown on the right.}
    \label{fig:qual-model}
\end{figure}

As in \cite{campbell_assessment_2021}, we hypothesize that certain species of fish show behavioral responses to vehicles, but that it is highly correlated with their sensing modalities. However, many of these quantities are difficult to measure and so, regardless of modality, there is some threshold distance at which their sensing or reaction is limited. It is known that many prey fish will shelter in the coral head or crevices if a threat is detected \cite{gonzalez-rivero_linking_2017, asunsolo-rivera_behaviour_2023}. We thus focus our study to estimating distances of fish to potential shelters, such as coral heads or other biological hotspots.

We propose a simple model of marine animal disturbance (hiding) based on a logistic function. Intuitively, as the AUV (or predator) approaches an animal, it will respond by moving closer to shelter (even if it means moving closer to the AUV). Our model is given by:

\begin{equation}
    f(x) = \frac{L_{control} - L_{hide}}{1 + e^{-k(x-x_0)}} + L_{hide}
    \label{eqn:model}
\end{equation}

Where $f(x)$ is predicted distance of fish from shelter. $L_{control}$ is average distance of fish from shelter assuming no disturbance, $L_{hide}$ is the same, but with disturbance. $x$ is the distance of the robot to the shelter (or the fish). $k$ and $x_0$ are constants that determine how reactive the fish are to disturbance. 

Biologists use a metric named \textit{flight initiation distance} (FID) \cite{asunsolo-rivera_behaviour_2023} as the distance when the flight response is first triggered. From our model above, we can directly compute the FID by setting $f(x_{FID}) = \alpha L_{control}$ and solving for $x_{FID}$, where $\alpha \in [0,1)$ depending on how conservative an estimate is desired. To mitigate disturbance, AUVs should attempt to operate outside the estimated FID. 

We note that in reality, many of these constants are likely functions of the environment, species, and the AUV itself. However, this also suggests that AUV designers can target reducing $k$ and $x_0$ to validate designs of new AUVs to minimize invasiveness.

\subsection{Field data acquisition}

\begin{figure}
    \centering
    \includegraphics[width=0.49\linewidth]{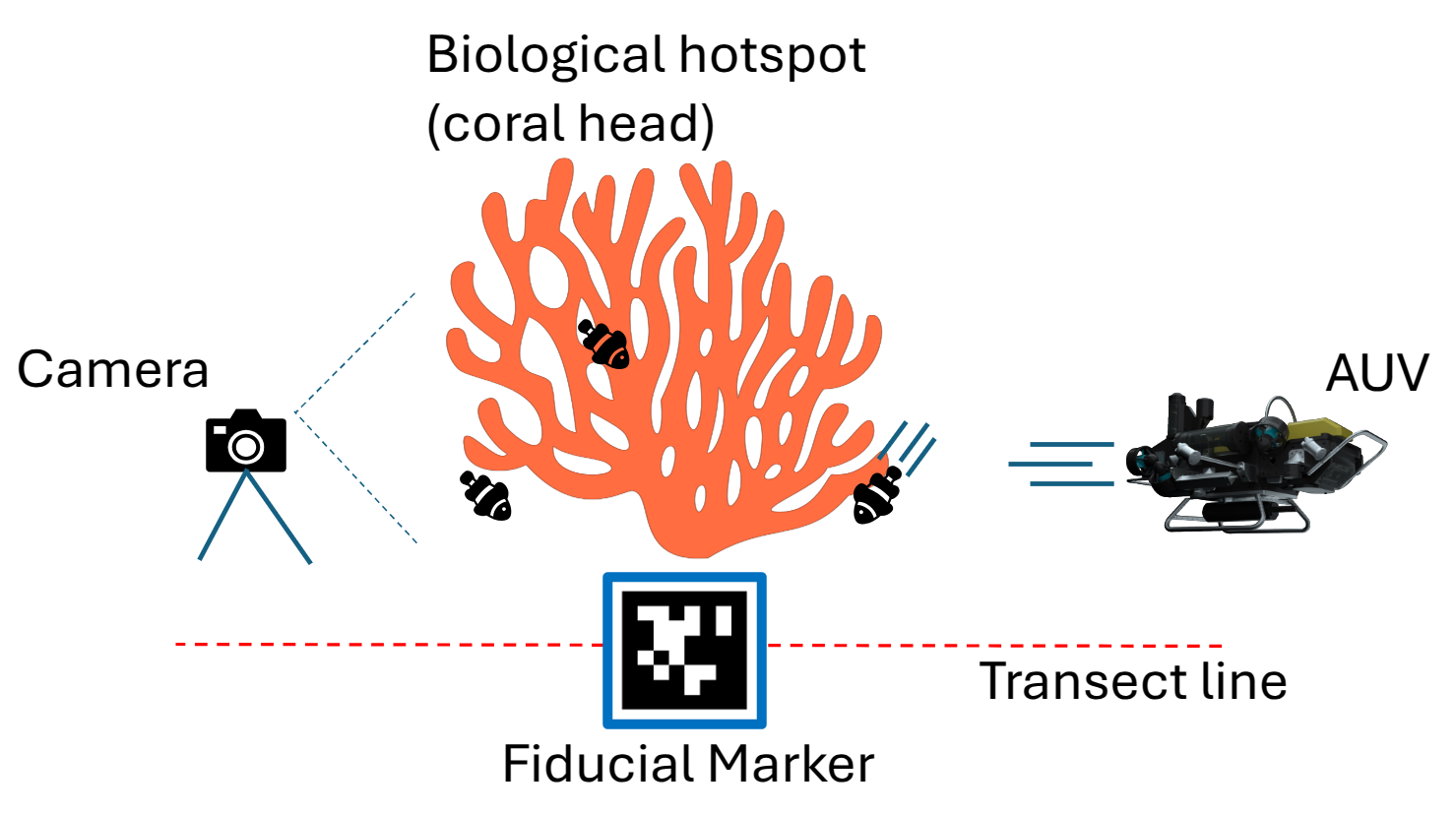}
    \includegraphics[width=0.49\linewidth]{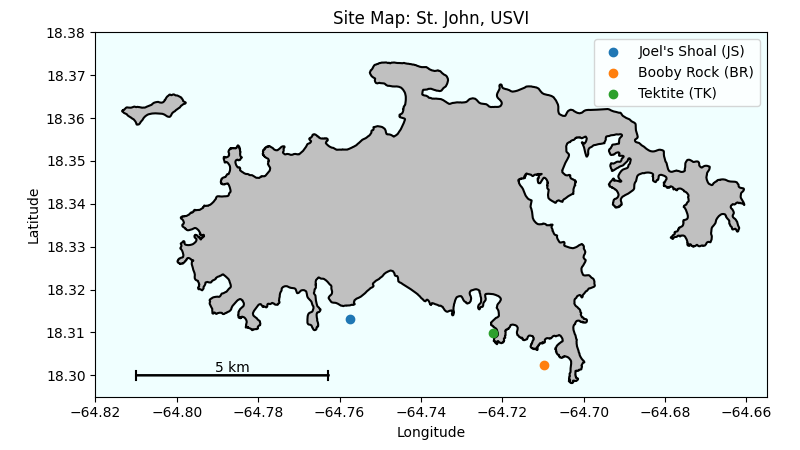}
    \caption{Left: Overview of our system for monitoring animal behavior in relation to a transiting underwater vehicle. The stationary camera is used to establish controls (animal behavior without underwater vehicles) as well as the interactions. 2 additional markers are placed 4 meters on either end of the coral marker to indicate transect stop and wait points. Right: Map of St. John, U.S. Virgin Islands with the coral reef sites marked where we conducted our field trials.}
    \label{fig:field-overview}
\end{figure}

Our data collection method is broadly illustrated in \Cref{fig:field-overview}. It used stationary cameras (GoPros) that were set to continuously capture video of a particular coral head, the shelter location, in a reef environment. We found it is crucial that the coral head have an established fish community in order to estimate control behaviors. We note that this limits the species that we can study with this methodology.

Next, we placed an AprilTag next to the coral head that was used for fine-grained localization of the AUV. We used a custom AUV/ROV \cite{girdhar_curee_2023} named CUREE, that utilizes 6 BlueROV T200 thrusters, to run several transects across the target coral head, though any AUV/ROV could be used as long as it is able to run consistent linear transects. After each pass of the AprilTag, the vehicle waited at least 1 minute in an idle state, roughly 4 m away from the AprilTag, in order to allow the fish behavior to re-acclimate. This time period was selected based on qualitative estimates from divers, but future work should study re-acclimation time periods and potential long-term acclimatization. We also recorded qualitative data that the fish did not acclimatize over the long-term, but further studies are needed to confirm this. Before the vehicle was placed in the water, we time-synchronized all the external cameras and vehicle on-board systems.

To test if distance is a variable of interest, we transited at various altitudes from the AprilTag and coral head. Importantly, these are variables that an AUV/ROV can control, and can form the basis for real-time feedback and mitigation efforts. We did collect data at various speeds, but analysis was more difficult, so will be part of future work.

\subsection{Data analysis}

We monitored fish behavior during periods where the vehicle passes over the AprilTag. For each transect, we extracted the image frame, from the synchronized stationary camera, where the vehicle was closest to the AprilTag. We manually annotated all fish seen in that frame in LabelBox. We also identified the pixel coordinate of the centroid of the biological shelter (in this case, coral head, usually associated with a \textit{Dendrogyra} in BR and JS, or the center of the coral head in TK). We also extracted 15-seconds of imagery on either side and for some videos, we also annotated frames from each 5 second interval. Next, we estimated the distances of each annotated fish to the shelter coordinate.

In-between each transect, the vehicle waited 4 m away from the AprilTag for at least 1 minute. Here, we extracted a single frame and annotated all fish in that image. These annotated images formed our control sets for each site. Similarly, we estimate fish distances to the shelter coordinate. In the future, we wish to automate this with trained fish detectors.


\section{Experiments}
\label{sec:experiments}


We collected data in the U.S. Virgin Islands, St. John, Virgin Islands National Park across multiple field campaigns. For this study, we use data collected in June 2023 and March 2024, from Booby Rock (BR), Joel's Shoal (JS) and Tektite (TK), shown in \Cref{fig:field-overview}. We conducted both diver and AUV transects. Sites were selected by divers several hours ahead of the transects, divers attempted to identify larger individual coral heads that seemed to have a consistent fish population. For the AUV transects, divers installed the cameras and AprilTag, and then waited for up to 10 minutes before initiating the transects. In total, we collected nearly 200 transects at 4 different coral reef sites (known as Booby Rock, Joel's Shoal, Tektite, and Yawzi reefs).

\begin{figure}
    \centering
    \includegraphics[width=\linewidth]{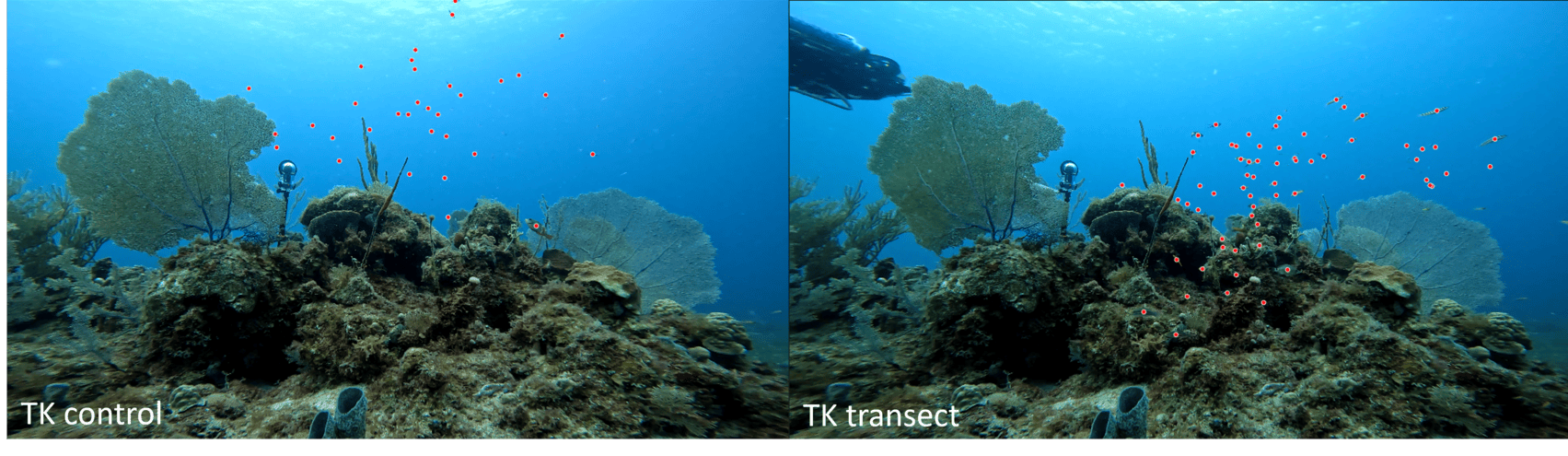}
    \caption[Example Tektite]{Here we show the control frame and an example transect frame and annotation at TK site, red dots indicate annotated fish.}
    \label{fig:qual-tk}
\end{figure}

The results for 46 transects, distributed across the 3 sites, and corresponding control time points (the vehicle hovers 4m away) are shown in \Cref{fig:qual-tk,fig:qual-multi-alt,fig:quant-disturbance}. In these cases, the vehicle was driven primarily manually in ROV mode from a moored or anchored boat on the surface. Many of the remaining transects were taken on days or hours when there was significant turbidity, under- or over-saturated environmental lighting conditions, or not a consistent fish population near the selected site, making them difficult to use. In all figures, the red dots are annotated fish locations.

\section{Results and Discussion}
\label{sec:discussion}

\subsection{Do fish respond to presence of underwater vehicles?}

\begin{figure}[ht]
    \centering
    \includegraphics[width=\linewidth]{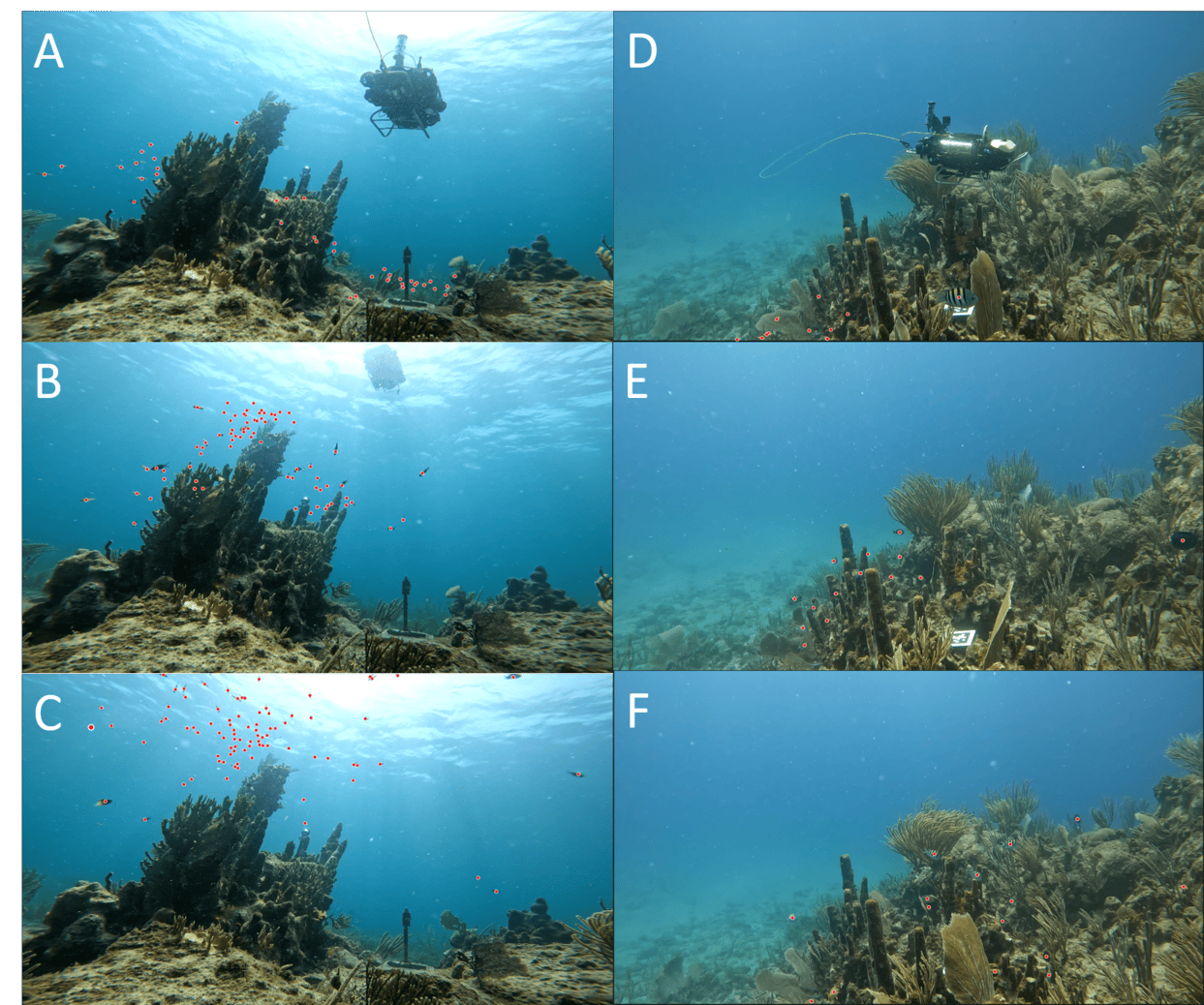}
    \caption[Multi-altitude Transects Qualitative]{Multi-altitude transects at BR (right) and JS (left) sites. Fish are annotated by red dots. Low-altitude transects are shown in A and D, high-altitude transects are shown in B and E, and controls are C and F. Most notably at BR (panels A-C), results suggest that there may be a sphere of influence of the vehicle, as the response is more muted for high-altitude transects (B) compared to low-altitude transects (A). The sparsity of fish in JS is more difficult to discern.}
    \label{fig:qual-multi-alt}
\end{figure}

\begin{figure}[h]
    \centering
    \includegraphics[width=0.49\linewidth]{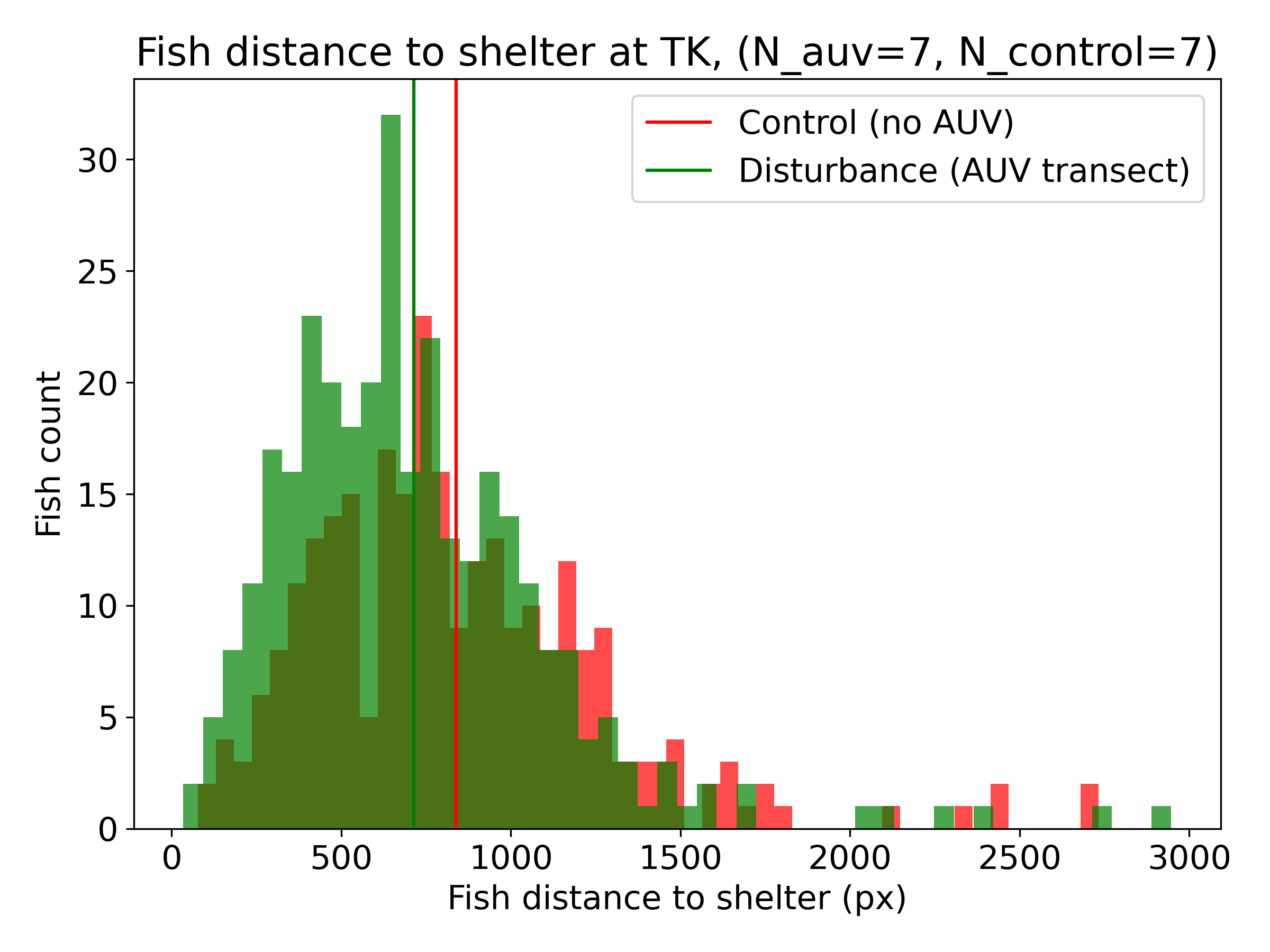}
    \includegraphics[width=0.49\linewidth]{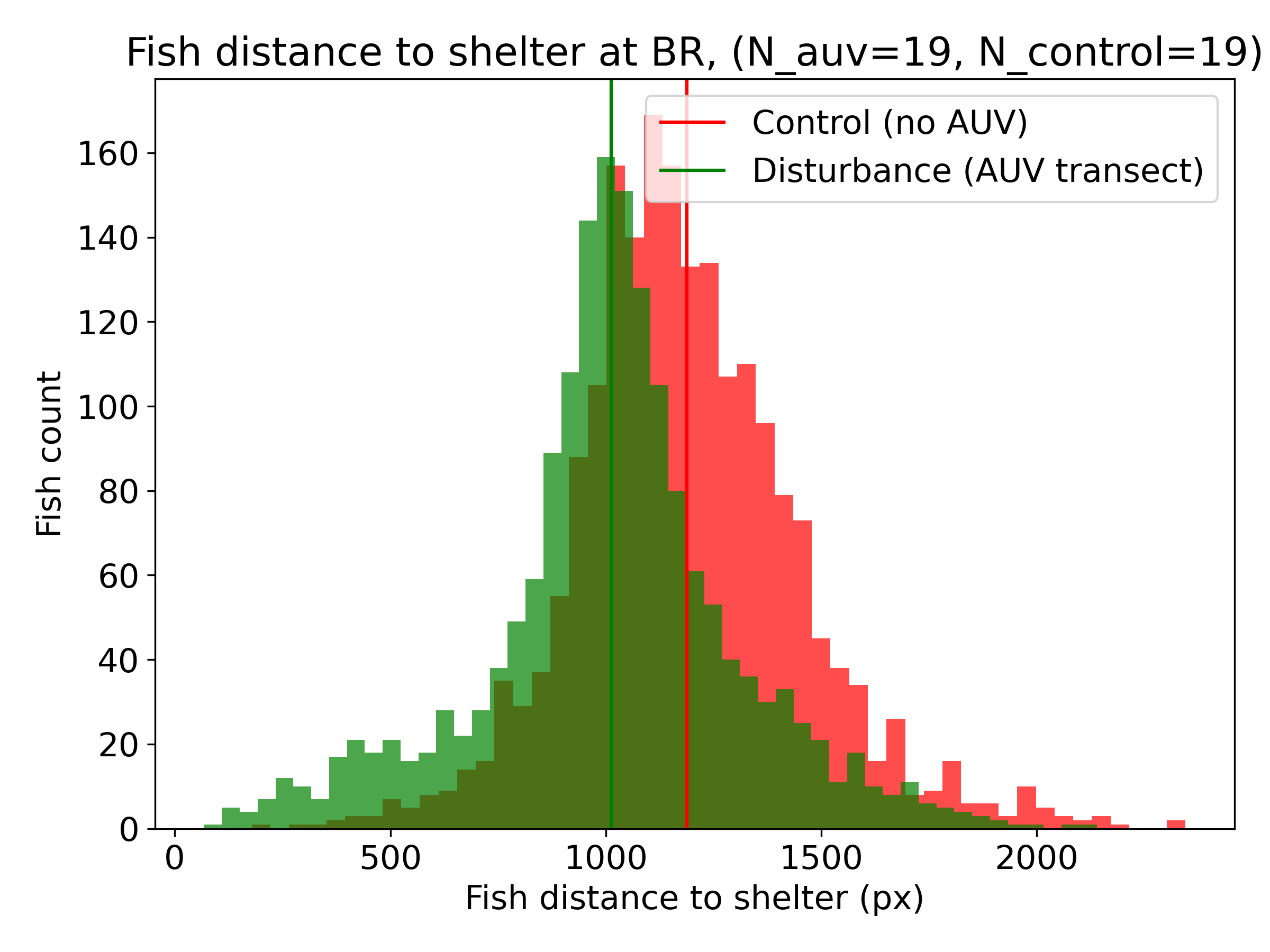}
    \includegraphics[width=0.49\linewidth]{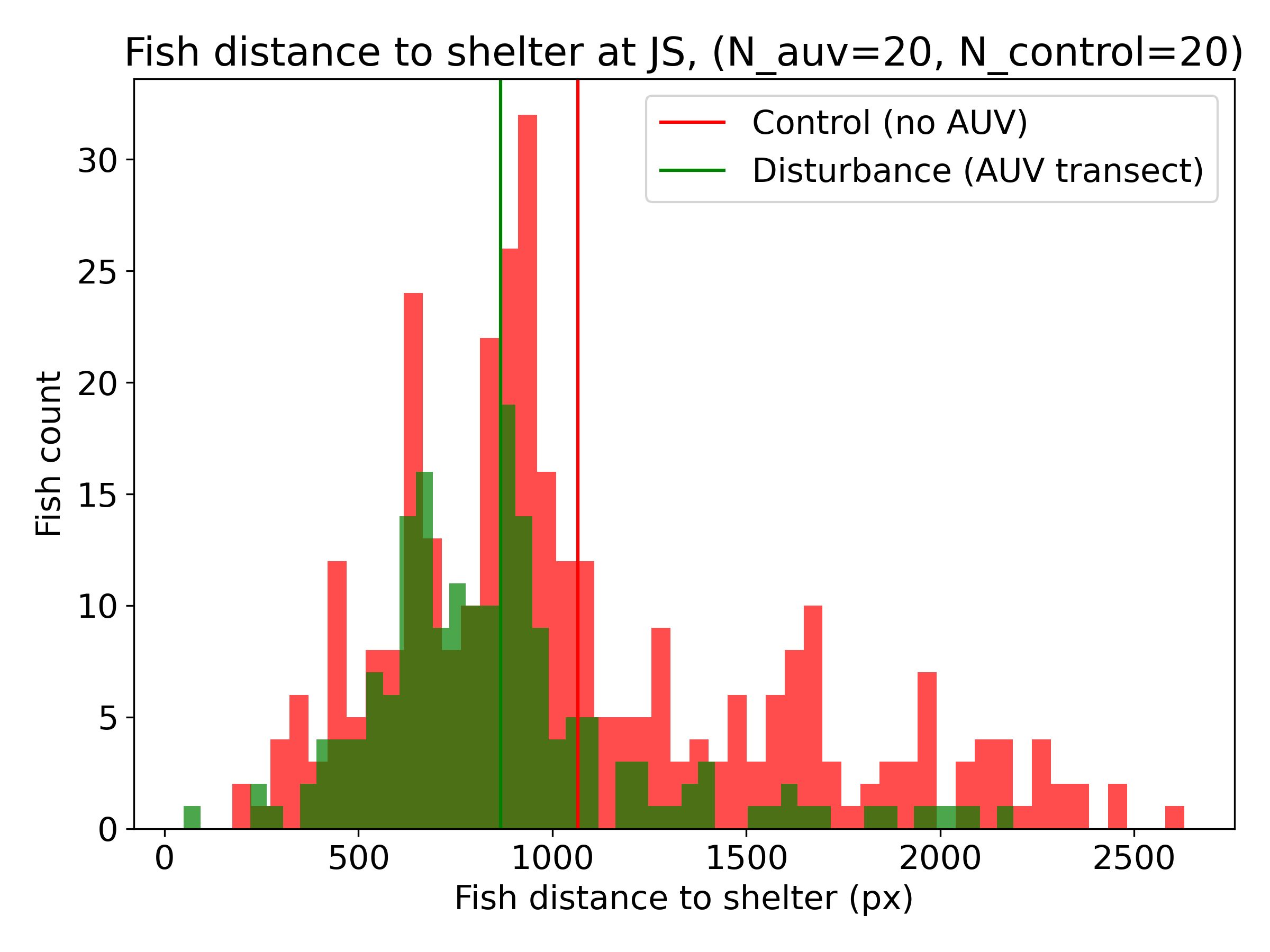}
    \caption[Fish disturbance at 3 sites]{Comparison of fish distance to coral reef hotspot in the absence (scientific control) and presence of AUV transects. The solid vertical lines are the means. N\_transect is the number of transects (1 frame labelled per transect) and N\_control is the number of control frames. Overall, there is a decrease in mean distance to the coralhead across all 3 sites, suggesting that AUV transects do trigger avoidance behaviors.}
    \label{fig:quant-disturbance}
\end{figure}

Qualitatively, in \Cref{fig:qual-tk,fig:qual-multi-alt}, we saw that some species of fish tend to stay further away from the coral sites by default. Most of the fish in these images are \textit{blue-headed wrasse} of various life stages, but particularly juveniles, and \textit{blue chromis}, though there are other species represented as well. However, during lower-altitude surveys, about 1.5m above the AprilTag, as in A and D in \Cref{fig:qual-multi-alt}, we saw a dramatic shift in the spatial distribution of the fish. In particular, they appeared to stay significantly closer to the coral head of interest, a \textit{dendrogyra} or pillar coral, as the vehicle approaches. This was especially apparent in A and C, at the BR site.

We performed quantitative analysis in \Cref{fig:quant-disturbance}, for each of the 3 sites, independently. There is a clear difference in the distribution of fish distances from the respective shelter points. When an AUV was transecting, the fish trended towards the shelter point. We ran a two-sample Kolmogorov–Smirnov test on each dataset, with the null-hypothesis being that the distributions are the same. We found p-values of 1.2e-4, 4.6e-70, and 3.1e-6 for TK, BR, and JS, respectively, suggesting that we reject the null hypothesis with confidence level of 95\%. Note that the vehicles altitudes were not standardized in these datasets.

Due to the use of monocular cameras, we were unable to estimate precise distances, but overall, we saw a stronger response at the BR site compared to the JS and TK sites. There are multiple potential explanations for this result, but one hypothesis is that the fish at TK are potentially more habituated to disturbances since it is a highly trafficked tourist site, comparatively. The JS fish population also seemed much smaller and more likely to stay near shelter by default, making it more noisy to estimate a large change in behavior.

These results suggest that the presence of the AUV caused fish to perform escape maneuvers towards shelter. Because they attempt to hide in the crevices of the coral head, this may reduce population estimates of these species of fish from the perspective of the vehicle. However, we note that we did \textit{not} see significant population differences from the stationary camera's perspective in these tests. Interestingly, from qualitative observations, \textit{Dendrogyra} corals tended to provide the most consistent field test regions.

\subsection{Are there strategies to mitigate this response?}

\begin{figure}
    \centering
    \includegraphics[width=0.49\linewidth]{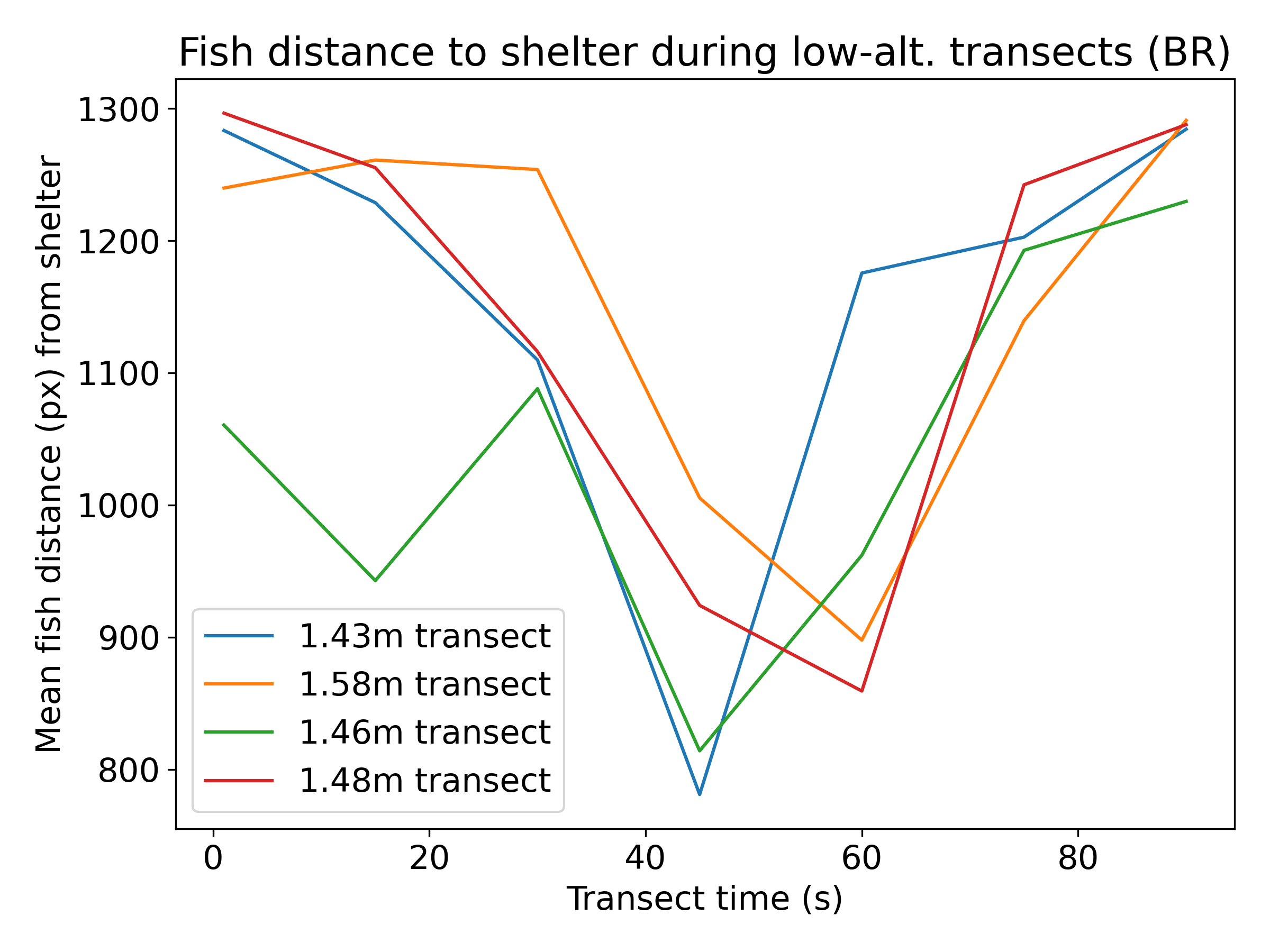}
    \includegraphics[width=0.49\linewidth]{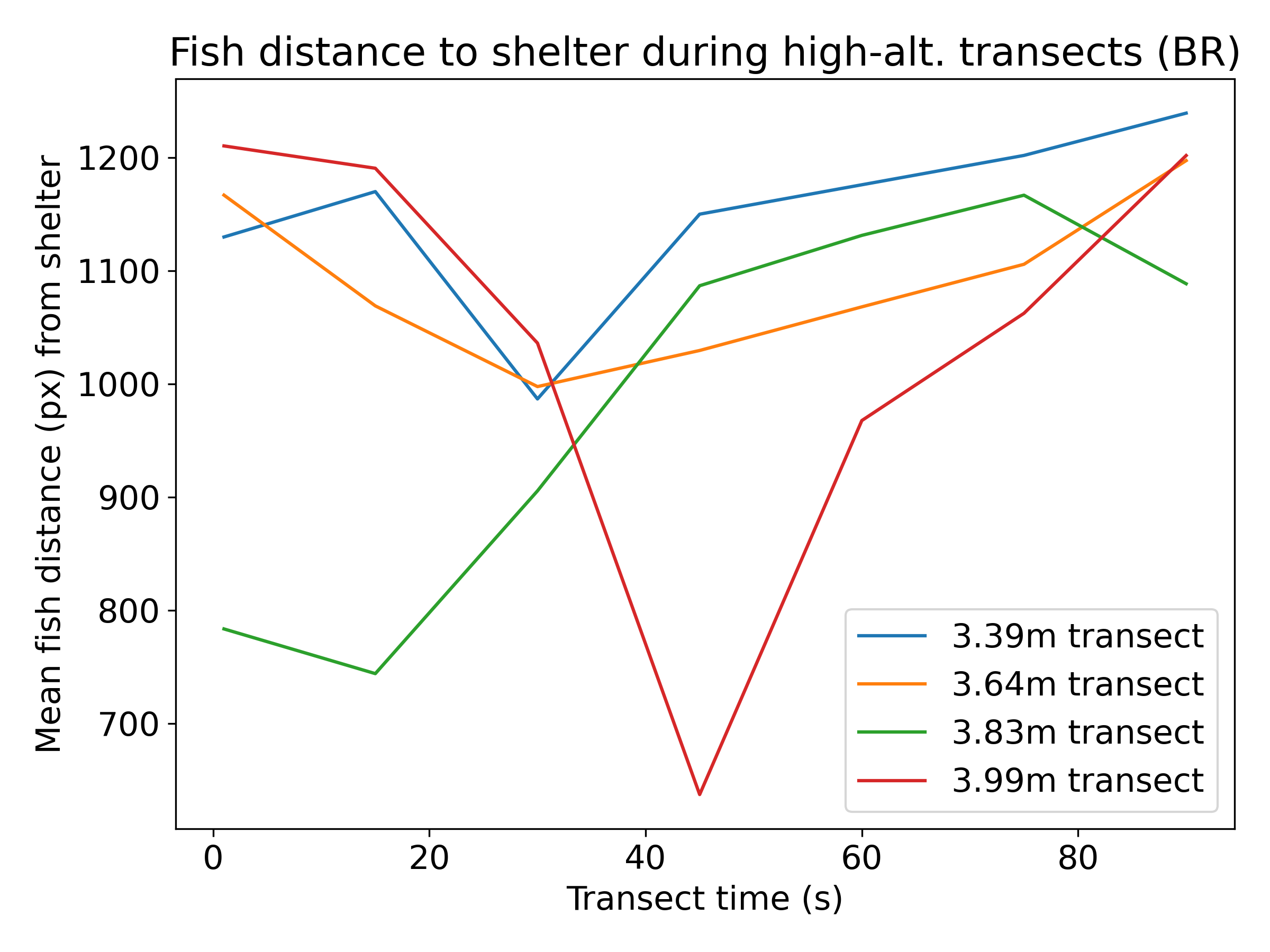}
    \includegraphics[width=0.49\linewidth]{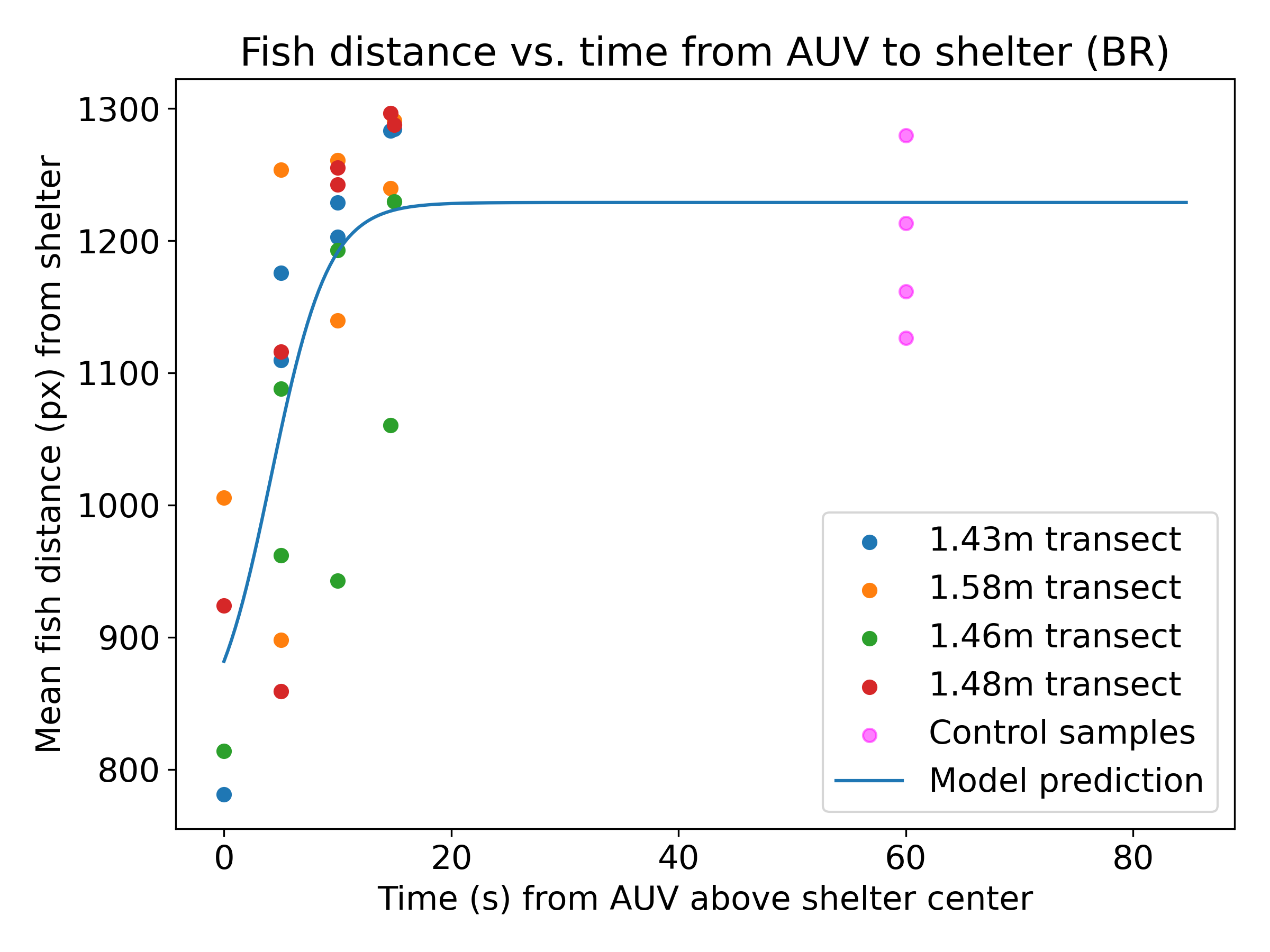}
    \includegraphics[width=0.49\linewidth]{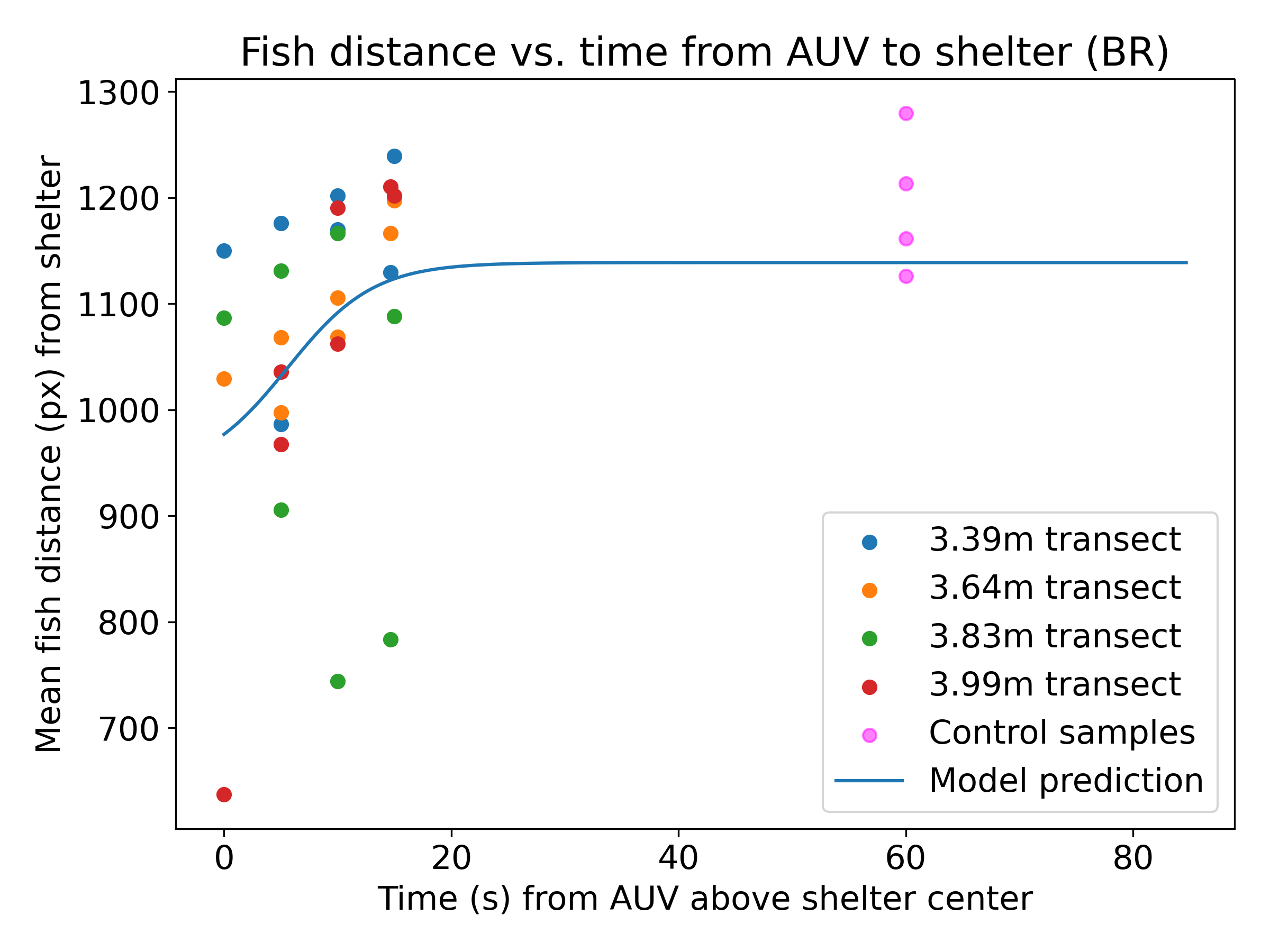}
    \includegraphics[width=0.49\linewidth]{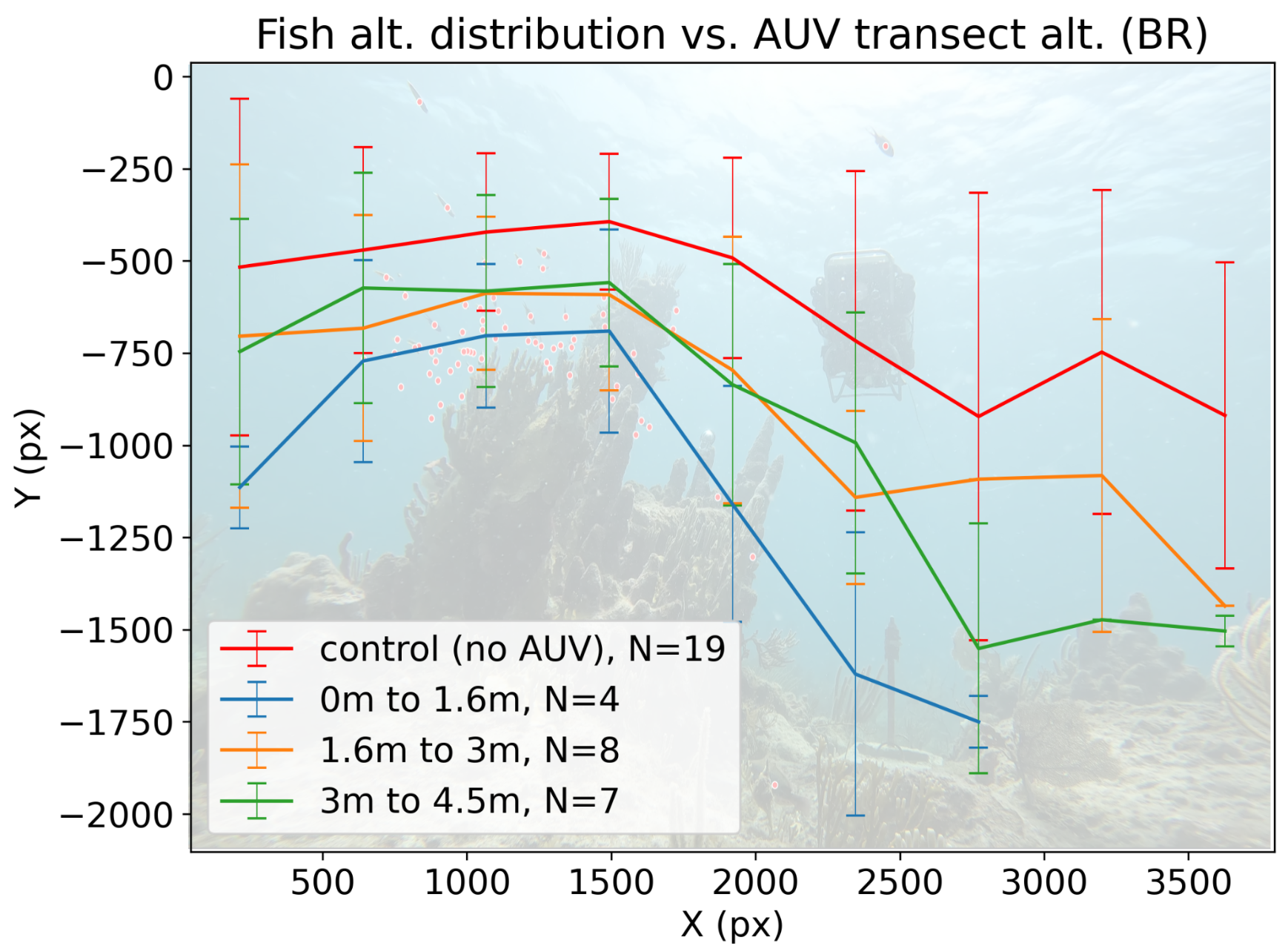}

    \caption{For site BR, we evaluate time-based (the top 4 plots) and transect altitude-based (bottom plot) analyses. Top row: each line  corresponds to the measured mean distance of fish to shelter during different transects. Middle row shows the same data from the plot directly above, but fit using our animal response model, with the corresponding control samples included (a control sample is taken every 60-seconds after the AUV passes the AprilTag). The bottom plot shows mean fish distances from shelter at the precise moment the AUV was directly above the AprilTag for 19 transects at various altitudes, demonstrating different behaviors from an altitude perspective.}
    \label{fig:quant-br-model-fit}
\end{figure}

One strategy for mitigating the impact of AUVs on fish response is to remain further away. The primary question is how steep is the response curve and is it possible for the vehicle to trade-off its own perceptual distance limitations with the response distance of the fish? Our study focused on the BR site where the response was most notable from the previous section.

We first investigated the response of fish between AUV transects at various altitudes, the analysis is shown in \Cref{fig:quant-br-model-fit} in the bottom panel. We see a general trend that at lower altitudes, fish tend to also move closer to the \textit{dendrogyra} shelter, especially if the AUV is transiting below 1.6m. Transects between 1.6m to 4.5m also showed some response (as the fish reside closer to shelter compared to the control state), but remain slightly further from the shelter than the low altitude transects. However, between 1.6m to 4.5m, there is not substantial difference.

As a second approach, we annotated timepoints \textit{along} each transect, shown in \Cref{fig:quant-br-model-fit} top panels. We used these to fit the model proposed in \Cref{eqn:model}, shown in \Cref{fig:quant-br-model-fit} middle panels. However, because of the data collection process, we were unable to extract a precise distance, and so use the time along the vehicle trajectory as a proxy for the vehicle distance. The vehicle was manually driven to maintain as constant a velocity as possible. Because control samples were taken exactly 60 sec after the AUV passes directly overhead an AprilTag, we used these control points to fit the time-based curve. Solving for FID using \Cref{eqn:model} (though in time rather than distance), we found FIDs of 12.99 sec for low-altitude and 16.19 sec for high-altitude transects. Interestingly, this suggests that fish hide earlier during high altitude transects (which can also be seen in the top right panel in \Cref{fig:quant-br-model-fit}), which may be because the vehicle is easier to see, but requires more data to validate. Additionally, because the vehicle was further, the gap between the measured $L_{control} - L_{hide}$ was smaller in the high-altitude transects, suggesting a muted response to the higher altitude transects. 

In general, this suggests it may be possible to adjust vehicle distance (or altitude) from fish (or coral) in order to minimize their impact on fish behavior and that the proposed model could be used to identify reasonable distances. At the moment, our results are preliminary. For future analysis, we hope to analyze the remaining videos in order to establish better statistical significance. Our data set also contains varying vehicle speeds to illustrate potential controls alternatives to altitudes, as well as comparison with human divers. Future experimental setups should use stereo-cameras to establish true distances and new strategies are needed to enable the study of animal-robot interactions for species that may not be nearby fixed coral heads.

\section{Conclusion}
This study provides initial evidence \textit{in-situ} that underwater vehicles used for monitoring marine animals, namely smaller coral reef fish, may impact their behavior. Furthermore, we provide a simple model of disturbance and data collection protocol that can be used to directly predict FIDs in the field, though it is currently limited to organisms that exhibit hiding behaviors and remain outside shelter normally. This FID value can inform vehicle behavior to reduce potential measurement bias by operating outside the predicted FID. The other model parameters can also be used to inform future AUV physical design decisions. Future work should provide direct distance metrics, establish further statistical significance, investigate sensing modalities (such as acoustics, etc.), and incorporate other species, behaviors, and environmental conditions. 

\section*{Acknowledgements}

Thanks to Nad\`ege Aoki for species identification, and WHOI Warplab and Mooney Lab, and the University of the Virgin Islands for field work assistance. The experiments were conducted under National Park Service Scientific Research and Collecting Permits VIIS-2022-SCI-0005 and funded by NSF grant \#2133029.


%
%
\printbibliography
\end{document}